\documentclass[a4paper, 10pt, conference]{ieeeconf}      

\IEEEoverridecommandlockouts                              

\usepackage[noadjust, nobreak]{cite}

\usepackage{graphicx}
\graphicspath{{./graphics/}}
\usepackage[utf8]{inputenc}
\usepackage{multirow}

\usepackage{xcolor}

\usepackage{amsmath}
\usepackage[noadjust]{cite}

\title{\LARGE \bf Towards Incorporating Contextual Knowledge into the Prediction of Driving Behavior}

\author{Florian~Wirthmüller\textsuperscript{\includegraphics[scale=0.4]{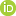}},
        Julian~Schlechtriemen\textsuperscript{\includegraphics[scale=0.4]{orcidlogo.png}},
        Jochen~Hipp\textsuperscript{\includegraphics[scale=0.4]{orcidlogo.png}}
        and~Manfred~Reichert\textsuperscript{\includegraphics[scale=0.4]{orcidlogo.png}}%
\thanks{F. Wirthmüller, J. Schlechtriemen and J. Hipp are with Mercedes-Benz AG, Böblingen, Germany, E-Mail: \{first\_name.last\_name\}@daimler.com.}
\thanks{F. Wirthmüller and M. Reichert are with the Institute of Databases and Information Systems (DBIS), Ulm University, Ulm, Germany,\newline E-Mail: \{first\_name.last\_name\}@uni-ulm.de}%
\thanks{J. Schlechtriemen is with the Institute of Realtime Learning Systems at the University of Siegen, Siegen, Germany.}
\thanks{ORCID: \href{https://orcid.org/0000-0002-9732-2561}{https://orcid.org/0000-0002-9732-2561};\newline \href{https://orcid.org/0000-0002-9130-061X}{https://orcid.org/0000-0002-9130-061X};\newline \href{https://orcid.org/0000-0002-9037-9899}{https://orcid.org/0000-0002-9037-9899};\newline \href{https://orcid.org/0000-0003-2536-4153}{https://orcid.org/0000-0003-2536-4153}}\thanks{\copyright~2020 IEEE. Personal use of this material is permitted. Permission from IEEE must be obtained for all other uses, in any current or future media, including reprinting/republishing this material for advertising or promotional purposes, creating new collective works, for resale or redistribution to servers or lists, or reuse of any copyrighted component of this work in other works.}
}

\usepackage{url}
\usepackage{hyperref}

\begin{document}
\IEEEoverridecommandlockouts
\pubid{\copyright~2020 IEEE}

\maketitle
\pagestyle{empty}


\begin{abstract}

Predicting the behavior of surrounding traffic participants is crucial for advanced driver assistance systems and autonomous driving. Most researchers however do not consider contextual knowledge when predicting vehicle motion. Extending former studies, we investigate how predictions are affected by external conditions. To do so, we categorize different kinds of contextual information and provide a carefully chosen definition as well as examples for external conditions. More precisely, we investigate how a state-of-the-art approach for lateral motion prediction is influenced by one selected external condition, namely the traffic density. Our investigations demonstrate that this kind of information is highly relevant in order to improve the performance of prediction algorithms. Therefore, this study constitutes the first step towards the integration of such information into automated vehicles. Moreover, our motion prediction approach is evaluated based on the public highD data set showing a maneuver prediction performance with areas under the \textit{ROC} curve above 97\,\% and a median lateral prediction error of only 0.18\,m on a prediction horizon of 5\,s.

\end{abstract}

\section{INTRODUCTION}

\begin{figure}[t!]
\centering\includegraphics[width=0.48\textwidth]{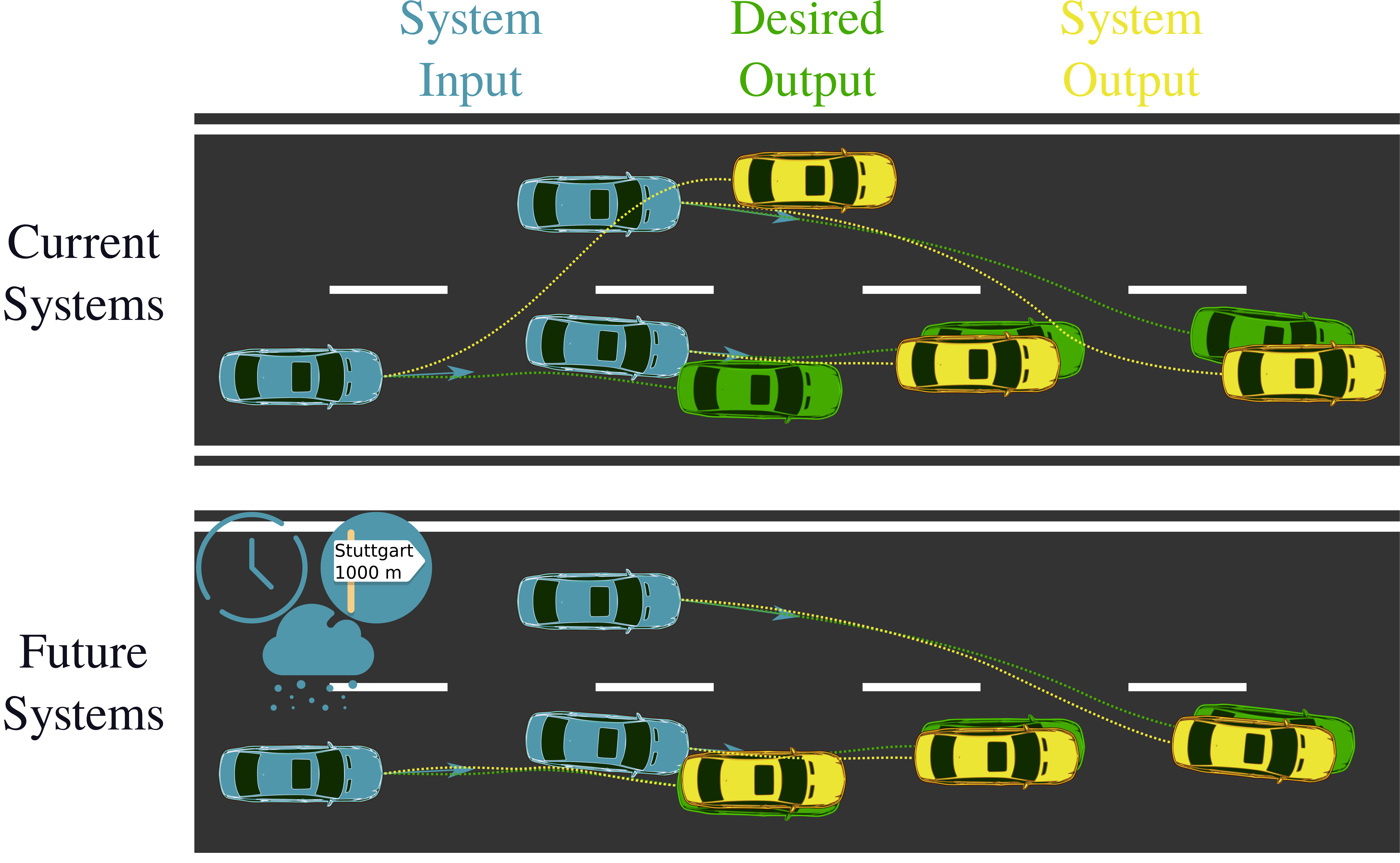}\caption{Most current motion prediction systems are not relying on external conditions as inputs (visualized in blue). Our hypothesis is that future systems can benefit by the integration of such information with an increased prediction performance. Therefore, the output (yellow) of a future system is in this example visualized as less optimistic and closer to the actual positions as desired output (green).}\label{fig:intro}
\end{figure}

When thinking about human driving behavior, it seems to be obvious, that it is not only affected by the current traffic situation, but  by various external conditions as well. For example, the weather situation, traffic density or daytime can depict such conditions. Knowledge about external conditions is also used by human drivers for improving their motion predictions of other traffic participants. This context-awareness is one important aspect distinguishing the ability of humans in predicting other vehicles movements from the one of current advanced driver assistance systems. Therefore, our hypothesis is that an improvement of the system's performance towards a human-like one can be achieved by taking contextual information and especially knowledge about external conditions into account when developing motion prediction systems. \autoref{fig:intro} visualizes this thought. In this paper, we are investigating the impacts of external conditions on driving behavior as well as on the performance of current motion prediction systems.

\autoref{subsec:problem} defines our understanding of the terms contextual information and discriminates between situation context and external conditions. Accordingly, we deduce the problem definition. \autoref{subsec:contribution} follows with the articles' contribution. Hence, \autoref{subsec:structure} closes the section with the following articles' structure.

\pubidadjcol

\subsection{Problem Definition}\label{subsec:problem}

We aim to investigate to which extent current motion prediction systems are affected by varying context situations and, thus, would potentially benefit from contextual information. Especially, we investigate the lateral motion behavior, which is more challenging to predict, but at the same time less investigated compared to longitudinal behavior. To do so, we have to provide a proper definition of the notion of \textit{context} first, as definitions and understandings are very heterogeneous \cite{baldauf2007survey}. If we look at research related to the term \textit{context-aware motion prediction}, most works use a context definition that solely includes features directly describing the traffic situation. Exemplary features include traffic rules \cite{gindele2013learning, klingelschmitt2015using}, intentions \cite{schneemann2016context}, information about the objects and lane markings within the scene \cite{gindele2013learning, bartoli2018context, pool2019context, messaoud2019relational, wirthmueller2019}, or map information as intersection distances or topologies \cite{gindele2013learning, pool2019context, lefevre2011context, lefevre2011exploiting, wiest2013incorporating}. In \cite{gindele2013learning}, all information on top of the vehicle kinematic is considered to be context. As this is a rather wide definition, including a lot of different aspects, we subdivide context features for motion prediction into two categories. The first category contains all above mentioned aspects, which are directly connected to the traffic situation and can be characterized as highly dynamic and mostly continuous. In the following, this type of context information is called \textit{situation context}. In contrast, we use the term \textit{external conditions} for features such as weather, daytime, traffic density or country, which apply to all vehicles in the given situation and can be characterized as quasi-static. Therefore, the problem we are tackling is to explicitly investigate the influence of external conditions on lateral driving behavior in contrast to other research focusing on the traffic situation context.

\subsection{Contribution}\label{subsec:contribution}

The contribution of this article is threefold:

\begin{enumerate}
\item We discuss and categorize contextual knowledge for motion prediction into \textit{traffic situation context} and \textit{external conditions}.
\item We adopt the prediction methods presented in \cite{wirthmueller2019} to a public data set and evaluate them carefully to enhance comparability.
\item We present examinations showing the impact of exemplary external conditions on driving behavior and, thus, on motion prediction. Particularly, we showcase the need for a fully-context-aware motion prediction approach.
\end{enumerate}

\subsection{Article Structure}\label{subsec:structure}

The remainder of this work is structured as follows: \autoref{sec:rel_work} discusses related work. \autoref{sec:data} summarizes the properties of the used data set and presents data preprocessing steps. Thus,  \autoref{sec:experiments} outlines our experiments and their results. Finally, \autoref{sec:conclusion} summarizes the contribution and gives an outlook on future work.

\section{RELATED WORK}\label{sec:rel_work}

This section gives an overview of related research on context-aware motion prediction. The term context is mostly used for what we call traffic situation context. The considered works include research focusing on maneuver estimation \cite{wirthmueller2019, lefevre2011context, lefevre2011exploiting} as well as works from the field of trajectory or position prediction \cite{gindele2013learning, klingelschmitt2015using, messaoud2019relational, wirthmueller2019, wiest2013incorporating}. Besides these approaches intended for vehicles, context-aware motion prediction is studied in other domains as well, such as human (e.g. \cite{schneemann2016context, bartoli2018context}) or cyclist  (e.g. \cite{pool2019context}) path prediction. Studies with regard to external conditions can be found in \cite{hoogendoorn2010longitudinal, hamdar2016weather}. These works investigate the impact of weather conditions on driving behavior but are not directly connected to the field of motion prediction.

\pagebreak

In \cite{lefevre2011context, lefevre2011exploiting}, an approach for predicting maneuvers at intersections using topological and geometrical characteristics is presented. The maneuver class is inferred with a Bayesian network incorporating uncertainties.

\cite{wiest2013incorporating} presents a Hierarchical Mixture of Experts (HME) architecture for predicting spatial probability distributions at intersections. The HME methodology divides the input space in a binary decision tree fashion. Thereby, in each sub-space the motion modeling is done by a specific expert. Within the division process contextual or categorical features can be incorporated as split criteria. The experts are infered from training data and are represented as conditional probability density function, which models the relationship between past and future motions. As a benefit of this approach the input features of the individual experts can remain constant, even if additional context features shall be added. The evaluation shows, that incorporating contextual information can be beneficial for the prediction performance and that a hierarchical division of the input space is preferable compared to an augmentation of the feature space. \cite{gindele2013learning} presents an approach for predicting positions at intersections based on a dynamic Bayesian network, combining modules learned from data and constructed by expert knowledge. \cite{klingelschmitt2015using} presents a probabilistic position prediction approach for road intersections, presuming a predefined maneuver estimation. The actual regression problem to find a tempo-spatial distribution is transformed by a discretization into a multiclass classification problem. The classification problem is addressed by training a neural net that, amongst others, contains map and traffic information as features.

Besides these approaches we want to mention two works, directly related to our studies. In \cite{messaoud2019relational}, for the first time, the highD data set (\cite{krajewski2018highd}), which is also used in this study, is used to implement and evaluate long-term motion predictions in highway situations. A relational recurrent neural network based encoder-decoder architecture is used for the predictions. The approach is able to predict lateral vehicle motions over a time horizon of 5\,s with a root mean squared error of 0.48\,m. Moreover, \cite{wirthmueller2019} presents a systematic machine learning workflow for the development of a system for maneuver detection and probabilistic motion prediction. It compares different classifiers and strategies, showing that a Mixture of Experts architecture using a multilayer perceptron classifier as gating node is beneficial compared to other combinations. This combination reaches a median lateral prediction error of 0.21\,m on a prediction horizon of 5\,s. To conduct our context-related studies, we use this architecture as starting point and will add further details in \autoref{subsec:setup}.

As aforementioned, there are only few works investigating the influence of external conditions on road traffic \cite{hoogendoorn2010longitudinal, hamdar2016weather}. These works study the impact of weather conditions on the driving behavior, but are not directly linked to motion prediction. \cite{hoogendoorn2010longitudinal} investigates longitudinal driving behavior under fog influence in a driving simulator study. The investigations show decreased speeds and accelerations as well as increased distances to preceding vehicles during foggy weather. \cite{hamdar2016weather} studies both, the impact of weather conditions and road geometries on longitudinal driving behavior when following a lead vehicle in a driving simulator study and gives a good literature survey in that area. Although \cite{hamdar2016weather} states that the overall effect of weather conditions is smaller than the one of challenging road geometries, the impact of weather conditions becomes apparent. As opposed to \cite{hoogendoorn2010longitudinal} the study shows that fog has only little impact on the driven speed, but in compliance with \cite{hoogendoorn2010longitudinal} fog implicates higher distances. In addition, lower speeds are observed when driving on wet or icy roads. 


Altogether, our literature review confirms, that context-aware prediction of driving behavior is studied by various works, most of them focusing on the traffic situation context, whereas external conditions have been neglected so far (cf. \autoref{subsec:problem}). As additional limitations, most works focus on urban driving and only investigate the behavior change on maneuver layer. However, as shown in \cite{hoogendoorn2010longitudinal, hamdar2016weather}, contextual features play also an important role for the concrete trajectory realization and not only for a maneuver decision. Interestingly, most works are studying longitudinal rather than lateral behavior.

\section{DATA PREPARATION}\label{sec:data}

To train and evaluate our algorithms, we rely on the recently published highD data set \cite{krajewski2018highd}. It consists of approximately 45\,000\,km of highway driving. The data set was recorded in 6 different highway sections in Germany using a drone-mounted camera system. As opposed to the formerly frequently used NGSIM data set \cite{colyar2007us}, the highD data set exhibits a considerably higher data quality and quantity as well as a higher variety.

We prepare the highD data set for our algorithms and context-related investigations, by performing several preprocessing steps: First, we calculate a range of additional features describing traffic density, lane changes and lane marking types as well as relations to the eight surrounding vehicles. The relations are described with a static environment model used by many researchers (see e.\,g. \cite{schlechtriemen2015will} for an introduction). Besides, we transform the data in an equal representation for both driving directions, containing the lane number in ascending order from right to left. In addition, we exclude data from the training and evaluation processes if front- or backsight distances are lower than 80\,m. This ensures that we only train and perform predictions based on a known system environment as expected in an in-vehicle application. In order to reflect real-world sensor capacities, we assume that vehicles are not able to perceive objects being farther away than 150\,m and put virtual sensor limitations by setting default values. Moreover, we adopted the strategy to label lane changes within a prediction horizon $T_{H}$ of 5\,s presented in \cite{wirthmueller2019}. As opposed to \cite{wirthmueller2019} we added a fourth label, representing samples with an undefined status $NDEF$ due to a too short observation time $T_{O}$. \autoref{eq:labeling} defines our labeling strategy.

$T_{LCL}$ and $T_{LCR}$ reflect the times to the next lane change to the left $LCL$ and to the right $LCR$, respectively. The $FLW$ label is used for lane following behavior. $T_{O}$ describes the remaining time meanwhile the object is in recording range and can thus be observed.

\begin{equation}
   L =
   \begin{cases}
     LCL,& \text{if } T_{LCL} \leq T_{H}\ \&\ \\ 
     & \; \; \; \;T_{LCL} \leq T_{LCR} \ \& \\
     & \; \; \; \;T_{LCL} \leq T_{O} \\
     LCR,& \text{if } T_{LCR} \leq T_{H}\ \&\ \\ 
     & \; \; \; \;T_{LCR} < T_{LCL} \ \& \\
     & \; \; \; \;T_{LCR} \leq T_{O} \\
     FLW,& \text{if } T_{H} < T_{LCL}\ \&\ \\ 
     & \; \; \; \;T_{H} < T_{LCR} \ \& \\
     & \; \; \; \;T_{H} \leq T_{O} \\
     NDEF,& \text{otherwise}\\
   \end{cases}
   \label{eq:labeling}
\end{equation}

After the labeling, we split all data with a defined maneuver class into 6 folds, of which 5 are used for training and the remaining one for evaluating the models. In our experiment, we only work with data from passenger cars, as the motion of cars and trucks deviates noticeable.

\section{EMPIRICAL STUDIES \& RESULTS}\label{sec:experiments}
This section first describes the experimental setup (\autoref{subsec:setup}). Afterwards, \autoref{subsec:results} presents empirical investigations with regard to the system's ability to predict upcoming maneuvers and future lateral positions. \autoref{subsec:results_context} presents investigations that show the impact of features associated with external conditions on the driving behavior. 

\subsection{Experimental Setup}\label{subsec:setup}

For performing our studies, we adopted the prediction approach we introduced in \cite{wirthmueller2019}. As aforementioned, our approach uses a multilayer perceptron to predict the upcoming maneuver. For the sake of simplicity, we re-used the hyper-parameters of the model and re-trained the maneuver classifier with the highD data set:

\begin{itemize}
	\item Step size: 0.02
	\item Structure: one hidden layer with 27 neurons; 3 output neurons
	\item Feature set: The considered feature set is corresponding to the one that had shown the best results in \cite{wirthmueller2019} except to the yaw angle, as this feature is not available in highD. Furthermore, a transformation to lane coordinates and a differentiation between the features in the different coordinate systems is not necessary in the given study, as applied in \cite{wirthmueller2019} as the highD data set solely contains straight road segments.
\end{itemize}

\begin{figure*}[ht!]
\centering\includegraphics[width=0.86\textwidth]{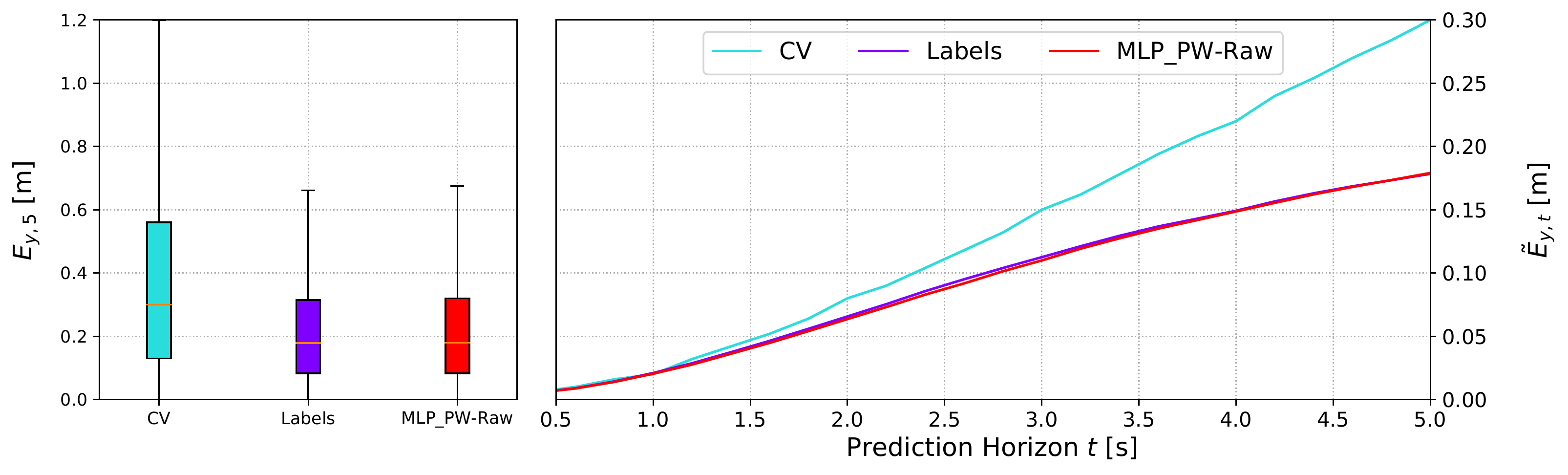}\caption{Distribution of the lateral error at t=5\,s (left) and the median lateral error as function of the prediction time (right).  The lateral error of the used system (denoted as 'MLP\_PW-Raw') is shown in comparison to a constant velocity ('CV') estimation and a Mixture of Experts assuming a perfect classifier ('Labels'). }\label{fig:error_boxplot}
\end{figure*}

In accordance with \cite{wirthmueller2019}, we used a random undersampling strategy to ensure that the training data set is balanced over time to the next lane change as well as over maneuver classes. 

Besides the multilayer perceptron, the approach uses three Gaussian mixture models, modeling future lateral positions depending on several input features. Thereby, each model is intended as expert for one of the three maneuver classes $LCL$, $FLW$ and $LCR$. To train the models, we again used the following hyper-parameters from \cite{wirthmueller2019} and retrained them with the highD data set variationally:

\begin{itemize}
	\item Maximum number of kernels: 50
	\item Type of the covariance matrix: full 
	\item Input features: lateral velocity $v_y$, distance to the center of the current lane $d_y^{cl}$ 
	\item Output dimensions: lateral position $y$, time $t$
\end{itemize}

The described individual components are used together in a Mixture of Experts fashion to predict distributions of future vehicle positions. A single position estimate can be calculated out of the distribution as center of gravity. For a more elaborate overview of the prediction technique and its parametrization see \cite{wirthmueller2019}.

\subsection{Prediction Performance Evaluation}\label{subsec:results}

\begin{figure}[h!]
\centering\includegraphics[width=0.40\textwidth]{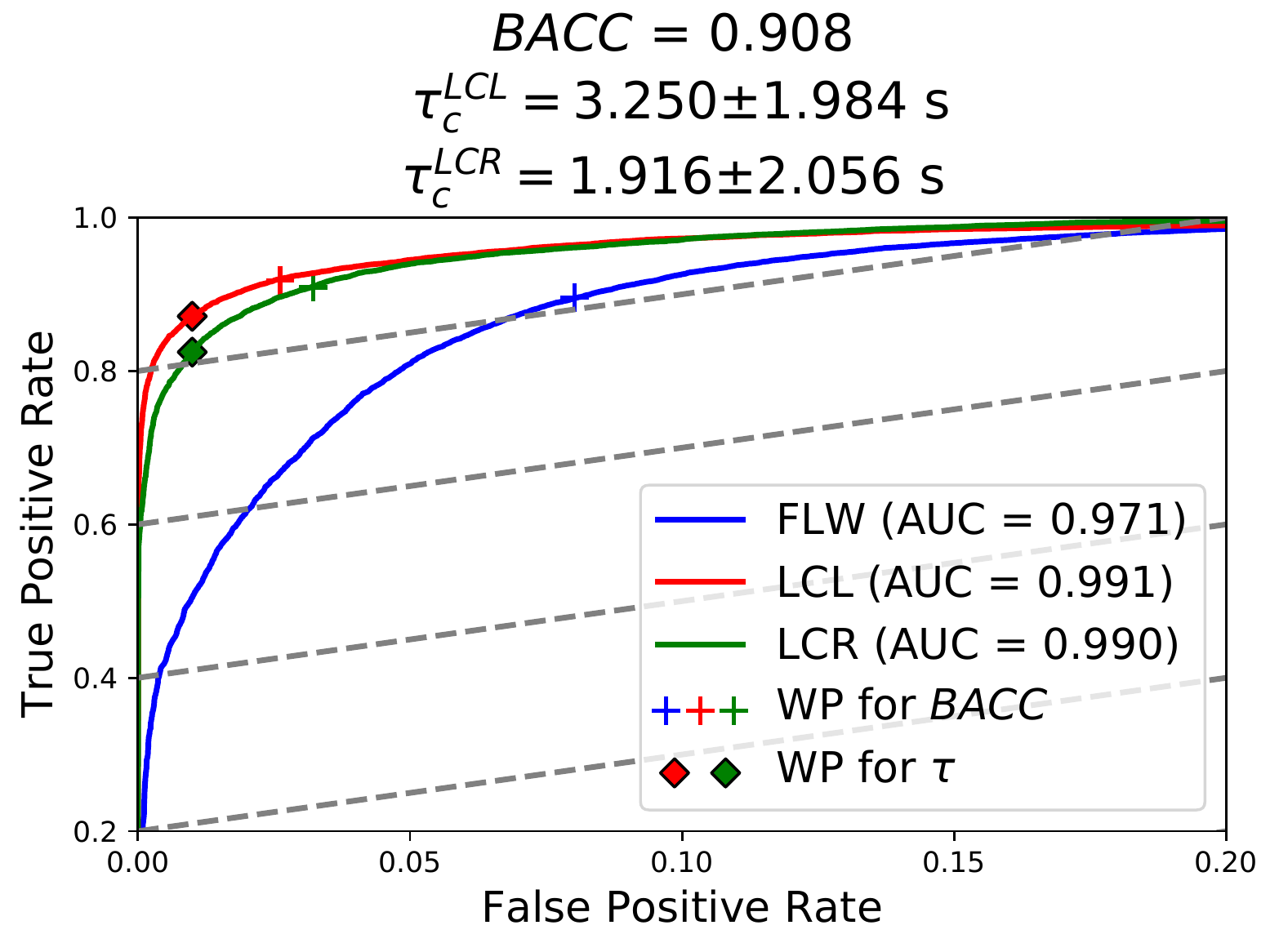}\caption{ROC curve showing the performance of the used maneuver classifier applied to the highD data set \cite{krajewski2018highd}.}\label{fig:ROC}
\end{figure}

To evaluate the performance of the maneuver classifier, we use the sixth data fold, which was left out during the models training. As evaluation metrics, we rely on the same measures as \cite{wirthmueller2019}: the balanced accuracy (\textit{BACC}),  the receiver operator characteristic (\textit{ROC}), the area under that curve (\textit{AUC}) and the time gain $\overline{\tau_c}$. Note, that  $\overline{\tau_c}$ measures the time between the vehicle center crosses the centerline and the moment from that the classifier is certain about its decision for a specific maneuver class and does not change it till the end of the situation. The results are presented in \autoref{fig:ROC}.

\begin{table}[!ht]
	\caption{Maneuver Classification Performance Compared to the Original Study \cite{wirthmueller2019}}
	\label{tab:ref_results}
	\centering
	\begin{tabular}{|c|c|c|c|c|c|}
		\hline
		& \multicolumn{3}{c|}{\textit{AUC}} & \multicolumn{2}{c|}{$\overline{\tau_c}$ [s]} \\
		& $LCL$ & $FLW$ & $LCR$ & $LCL$ & $LCR$ \\
		\hline
		\cite{wirthmueller2019} & 0.976 & 0.915 & 0.960 & 2.72 & 2.68 \\
		\hline
		This study & 0.991 & 0.971 & 0.990 & 3.250 & 1.916 \\
		\hline
		\hline
		Benefit & +0.015 & +0.066 & +0.030 & +0.530 & -0.764 \\
		\hline 
	\end{tabular}
\end{table}

As \autoref{tab:ref_results} shows, the maneuver classification performance is even superior to the results presented in \cite{wirthmueller2019}. This can be explained with the fact that the highD data set does not contain curved roads, as in \cite{wirthmueller2019} where an error-prone and complex transformation to curvilinear coordinates had to be performed. Solely the time gain for lane changes to the right drops, although the overall maneuver classification performance increases. This may be explained with the removal of the yaw angle from the feature set, resulting in a decreased classification stability.

For evaluating the performance of the position prediction, we use the distribution of the lateral prediction error $E_{y,\,5}$ at a prediction time of 5\,s and the median lateral error $\tilde{E}_{y}$ as a function of the prediction time $t$ as metrics. The results of our approach are visualized in comparison to a constant velocity (CV) prediction and a Mixture of Experts assuming a perfect maneuver classification (Labels) in \autoref{fig:error_boxplot}. 

As the evaluation shows, the maneuver classifier seems to be that good, that the performance of the down-streamed position prediction is nearly as good as when presuming a perfect classifier. When looking at the errors as a function of the prediction horizon, it becomes apparent that at some prediction horizons (e.\,g. around 2.5\,s) our approach is even better than a perfect classifier. This can be explained with the fact that the predicted position is calculated as a correctly weighted mixture of different hypotheses. As opposed to the original study \cite{wirthmueller2019}, the median lateral position error $\tilde{E}_{y,\,5}$ at a prediction horizon of 5\,s is decreased from less than 0.21\,m to less than 0.18\,m.

\begin{figure}[t!]
\centering\includegraphics[height=5.8cm]{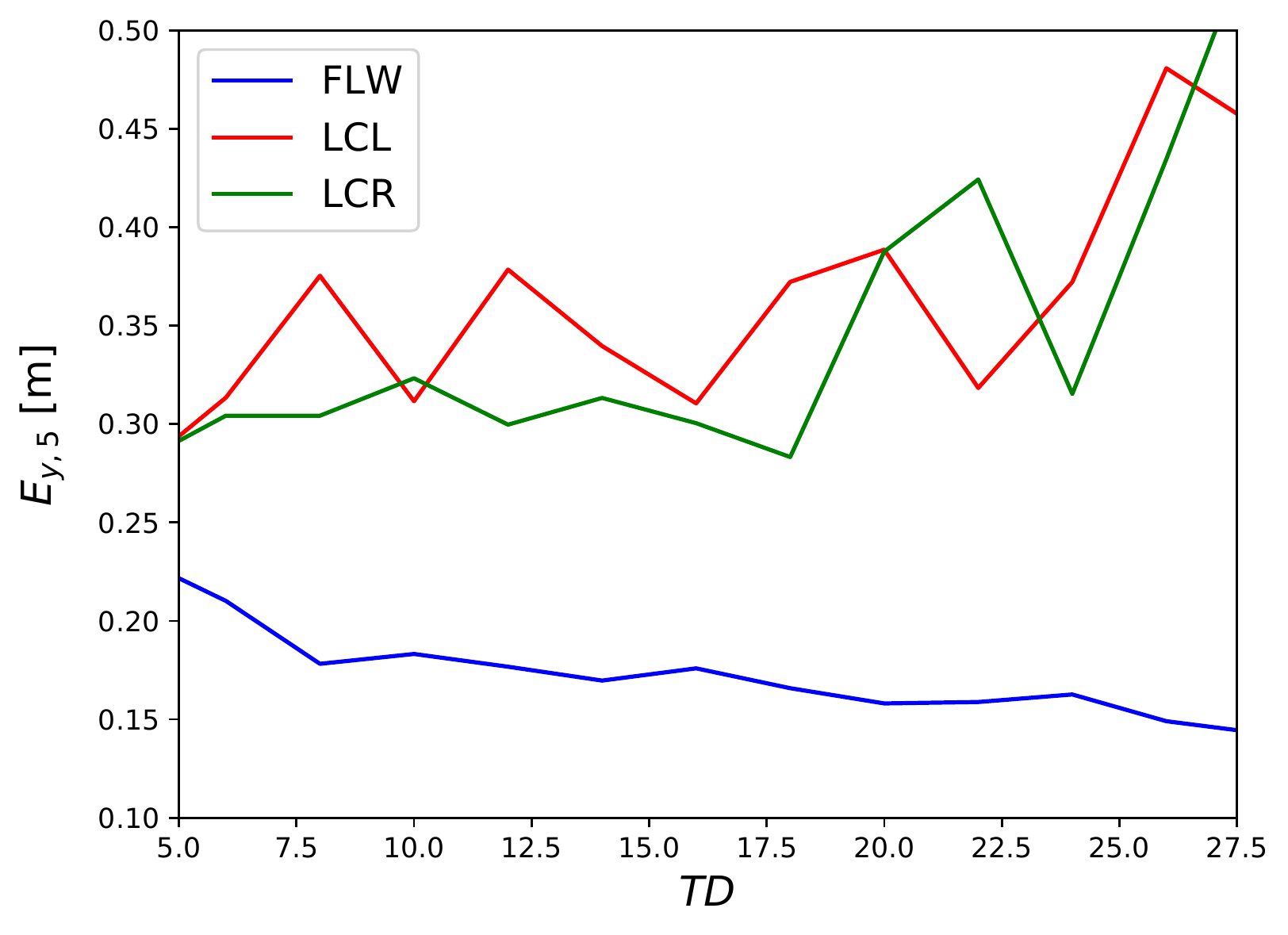}\caption{Median lateral prediction error on a prediction horizon of 5\,s as function of the traffic density for the different maneuver classes.}\label{fig:error_over_td}
\end{figure}

\subsection{Context Influence Investigations}\label{subsec:results_context}

As discussed, our main goal is to investigate and substantiate  the impact of features associated with external conditions on lateral driving behavior predictions. To accomplish this goal, first, we need to discuss which conditions can be investigated appropriately based on the available data set:


\subsubsection{Daytime}\label{subsec:daytime}
Although the highD data set contains daytime data, they are limited to a range from 9\,am to 8\,pm. Thus, the nighttime, for which we would expect a significant behavior change due to poor visibility conditions, is not included. We further believe that other effects, such as commuters, can be preferably represented with other features such as traffic density.


\subsubsection{Day of the Week}\label{subsec:weekday}
As the highD data set does not contain recordings at weekends and the data density over the other days of the week is very unbalanced, we do not consider investigations concerning that property to be fruitful.


\subsubsection{Weather}\label{subsec:weather}
Other researchers as e.\,g. \cite{hoogendoorn2010longitudinal, hamdar2016weather} already showed that the impact of varying weather conditions (on the road e.\,g. icy or wet roads, or above the road e.\,g. rain, fog or snow) on driving behavior is remarkable. As the highD data set, however, is recorded with optical drone-mounted sensors, sufficient visibility as well as flight conditions have to be satisfied. Consequently, the data set does not contain such data points.

\subsubsection{Location}\label{subsec:location}
Another external condition constitutes the location, which has assorted characteristics as well. For example, the context \textit{location} may depict the current country, prevalent speed limits, special road elements (e.\,g. bridges, tunnels or on-/off-ramps), patterns describing the road geometry or only one particular geo-location. As most other mentioned characteristics of the location context are hard to represent and to investigate due to an insufficient data situation, we are very optimistic with respect to the investigation concerning different countries. While the highD data set, as of now, only contains data recorded on German highways, we aim to gather additional data in the same format from other countries.


\subsubsection{Traffic Density}\label{subsec:traffic_density}
Another external condition whose impact on driving behavior is apparently obvious constitutes the traffic density. As opposed to other effects, however, the traffic density offers two important benefits. In the first place, we are able to calculate traffic density measures for the highD data set in a very stable way due to the vertical view onto the scene. In addition, we are able to represent also other effects as previously mentioned in the discussion concerning daytimes through the traffic density. Therefore, we believe that the most promising research direction, is to investigate the influence of the traffic density. 



To work with that property, we calculate the density values $TD$ in the highD data set according to \autoref{eq:traffic_density} as number of vehicles $N^V$ per km and number of lanes $N^L$ (cf. \cite{begriffsbestimmungen2000verkehrsplanung}):
\begin{equation}
   TD = \frac{N^V}{km \cdot N^L}
   \label{eq:traffic_density}
\end{equation}

\autoref{fig:error_over_td} presents the median lateral prediction error $\tilde{E}_{y,\,5}$ on a prediction horizon of 5\,s  as function of the traffic density for the different maneuver classes. This investigation shows contrary trends for lane change and lane following situations. While the prediction error during lane following decreases with increasing traffic density, it increases or at least remains in the same magnitude during lane changes.

\begin{figure}[t!]
\centering\includegraphics[height=5.8cm]{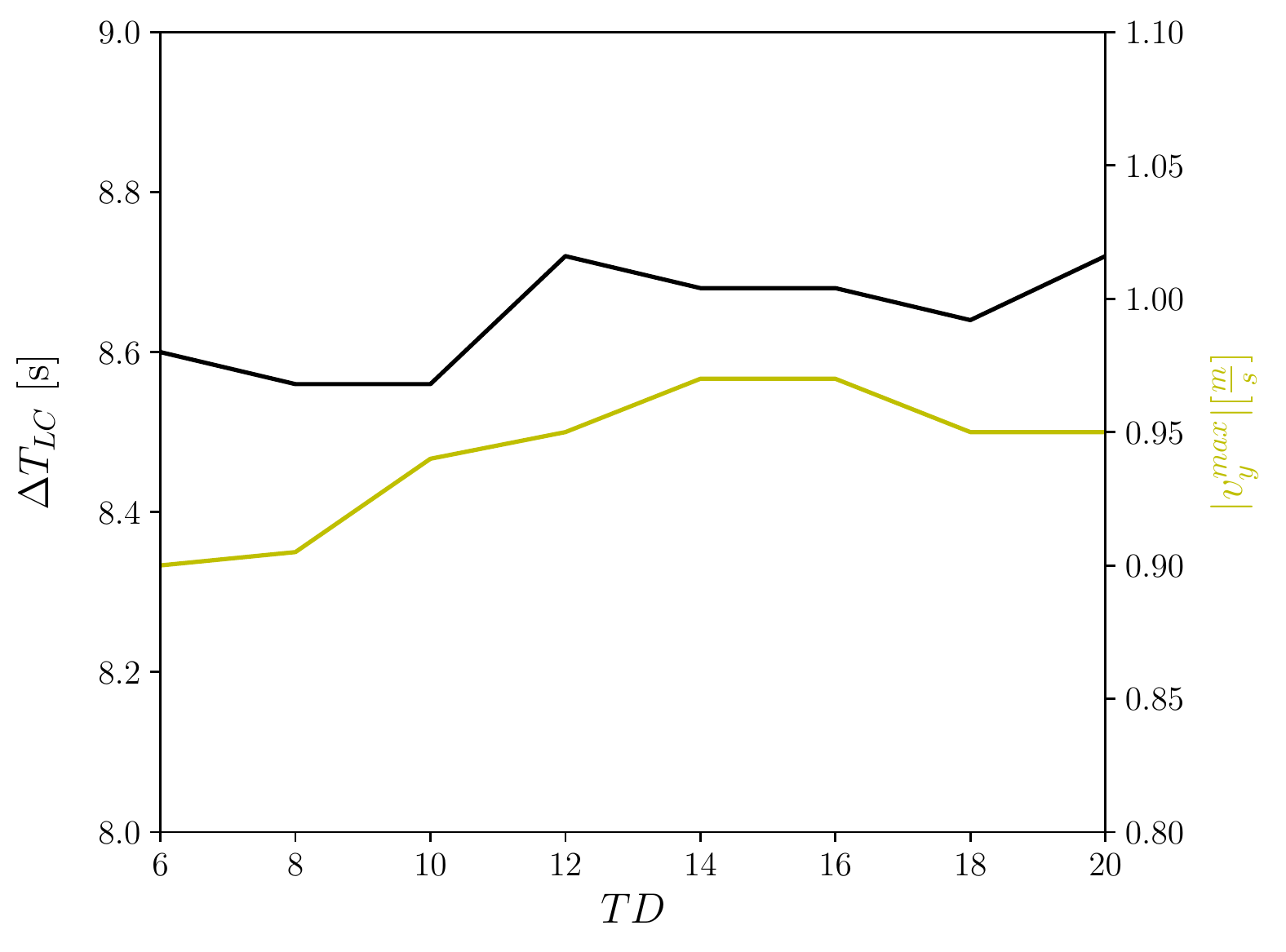}\caption{Lane change duration (black) and maximum lateral speed (yellow) during lane change as functions of the traffic density}\label{fig:lane_change_duration}
\end{figure}

A possible explanation for this observation could be that the increased traffic density decreases the average longitudinal speed. On the one hand, the lateral oscillation, human drivers normally conduct when following a highway lane, has a smaller amplitude in the local area, as the  frequency is constant. On the other, the duration of lane changes increases while the traffic becomes more dense. For example, that effect is substantiated in \cite{toledo2007modeling} and confirmed by our own investigations in \autoref{fig:lane_change_duration}. 

\begin{figure*}[ht!]
\centering\includegraphics[width=0.99\textwidth]{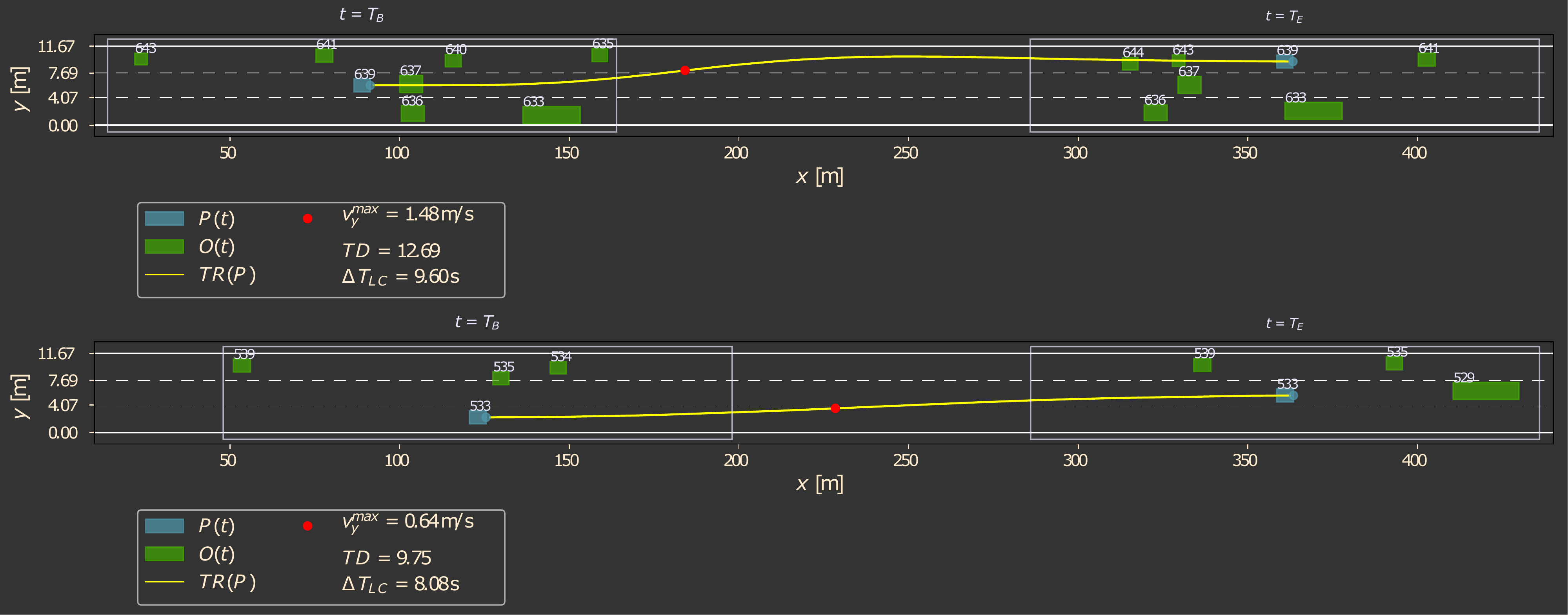}\caption{Visualization of two lane change situations out of the highD data set (recording 35). Whereas, the lane change in the upper example is performed under an increased traffic density and abruptly with a high maximum lateral velocity,  the one in the lower example is performed very smoothly. The visualization shows for both examples the positions of the prediction target $P(t)$ and of the relevant surrounding objects $O(t)$ at the begining $T_B$ and the end $T_E$ of the lane change maneuver as well as the whole trajectory of the prediction target $TR(P)$.}\label{fig:situations}
\end{figure*}

Moreover, \autoref{fig:lane_change_duration} shows that the maximum lateral speed during lane changes tends to increase, while the traffic density increases. This means, that one can imagine the procedure of a lane change during dense traffic as a long period of time where the driver slightly steers towards the lane marking and announces his lane change intention. Afterwards, the actual lane change is performed very abrupt and with a high maximum lateral speed, if a proper gap is reachable. In contrast, in light traffic, lane changes are performed slower and smoothly. To illustrate this, \autoref{fig:situations} shows two lane change maneuvers from the highD data set. As result of the described effects, it becomes presumably more complex to predict lane changes in more dense traffic as the variance increases.


To investigate this, we defined the duration of lane changes $\Delta T_{LC}$ according to \autoref{eq:lc_duration} as time difference between begin $T_{B}$ and end $T_{E}$ of the lane change:

\begin{equation}
  \Delta T_{LC} = T_ {E} - T_{B}
   \label{eq:lc_duration}
\end{equation}

$T_{B}$ and $T_{E}$ are defined as the first points in time before and after the lane change satisfying the conditions $C(T_{B})$ and $C(T_{E})$ according to \autoref{eq:t_beg} and \autoref{eq:t_end}:

\begin{equation}
\begin{aligned}
  C(T_{B}) = |d^{cl}_y| \geq 1\,m\ \\
 \vee |v_y| \geq 0.1\,\frac{m}{s}\\
 \vee |a_y| \geq 0.1\,\frac{m}{s^2}\\
 \vee |j_y| \geq 0.1\,\frac{m}{s^3}
   \label{eq:t_beg}
\end{aligned}
\end{equation}

\begin{equation}
\begin{aligned}
  C(T_{E}) = \overline{C(T_{B})}
   \label{eq:t_end}
\end{aligned}
\end{equation}


$d^{cl}_y$ describes the lateral distance to the lane center, $v_y$ the lateral velocity, $a_y$ the lateral acceleration and $j_y$ the lateral jerk. We determined the actual threshold values through visual inspection of all lane change situations.

\section{SUMMARY AND OUTLOOK}\label{sec:conclusion}

Other works have already shown the impact of external conditions such as weather (\cite{hoogendoorn2010longitudinal, hamdar2016weather}) on driving behavior. We are complementing this with investigations with respect to the traffic density. Our study goes beyond others, not only examining the impact on driving behavior itself, but instead the direct impact on the prediction performance. The results substantiate that motion prediction systems for automated vehicles could significantly benefit from explicitly considering what we call external conditions.


Besides, we investigate the prediction performance of the approach originally presented in \cite{wirthmueller2019} with the publicly available highD data set. The examinations show a maneuver classification with areas under the curve above 97\,\% and a median lateral prediction error of less than 0.18\,m for a prediction horizon of 5\,s.

As extension of the presented study, a full adaption of the approach presented in \cite{wirthmueller2019}, including the entire featureselection and hyper-parameter optimization pipeline, could further improve the results. In addition, we are preparing a follow-up study, presenting a fleet-learning-based architectural concept. Thereby, we will show how relevant data for context-aware motion prediction applications can be collected and used to ensure a continuous improvement of a vehicle fleets prediction capabilities under varying contextual influences. To enable more context-related investigations it is also desirable to conduct measurements during various contexts as night, rain, snow or in different countries.

\addtolength{\textheight}{-14.3cm}   

\bibliographystyle{ieeetr}

\bibliography{bib_itsc2020}

%
%
%

\end{document}